%% file: main.tex
\pdfoutput=1

\documentclass[11pt]{article}

\usepackage{acl}

\usepackage{times}
\usepackage{latexsym}

\usepackage[T1]{fontenc}

\usepackage[utf8]{inputenc}

\usepackage{microtype}

\usepackage[ruled,vlined]{algorithm2e}
\usepackage{graphicx}
\usepackage{wrapfig}
\usepackage{amsmath,bm,bbm,upgreek}
\usepackage{amsthm}
\usepackage{xcolor}
\usepackage{caption}
\usepackage{subcaption}
\usepackage{floatrow}
\usepackage{url}
\usepackage{booktabs}
\usepackage{multirow}
\usepackage{wrapfig}
\usepackage{floatflt}
\usepackage{lipsum}
\usepackage{makecell}
\usepackage{mathtools}
\usepackage{dblfloatfix}

\newcommand{\ure}[1]{#1}

\usepackage{easyReview}

\title{Towards Large-Scale Interpretable Knowledge Graph Reasoning for Dialogue Systems}

\author{
    Yi-Lin Tuan$^1$, Sajjad Beygi$^2$, Maryam Fazel-Zarandi$^2$\\
    {\bf Qiaozi Gao$^2$, Alessandra Cervone$^2$, William Yang Wang$^1$}\\
    $^1$ University of California, Santa Barbara\quad
    $^2$ Amazon Alexa AI\\
    \texttt{\{ytuan, william\}@cs.ucsb.edu}\\
    \texttt{\{beygi, fazelzar, qzgao, cervon\}@amazon.com}\\
}

\begin{document}
\maketitle
\begin{abstract}

Users interacting with voice assistants today need to phrase their requests in a very specific manner to elicit an appropriate response.
This limits the user experience, and is partly due to the lack of reasoning capabilities of dialogue platforms and the hand-crafted rules that require extensive labor.
One possible \ure{way} to improve user experience and relieve the manual efforts of designers is to build an end-to-end dialogue system that can do reasoning itself while perceiving user's utterances.
In this work, we propose a novel method to incorporate the knowledge reasoning capability into dialogue systems in a more scalable and generalizable manner.
Our proposed method allows a single transformer model to directly walk on a large-scale knowledge graph to generate responses. 
To the best of our knowledge, this is the first work to have transformer models generate responses by reasoning over differentiable knowledge graphs. We investigate the reasoning abilities of the proposed method on both task-oriented and domain-specific chit-chat dialogues.
Empirical results show that this method can effectively and efficiently incorporate a knowledge graph into a dialogue system with fully-interpretable reasoning paths.

\end{abstract}

\input{sections/01introduction}

\input{sections/02related_work}

\input{sections/03background}

\input{sections/04method}

\input{sections/05experiments}

\input{sections/06analysis}

\input{sections/07conclusion}

\bibliography{anthology,custom,main}
\bibliographystyle{acl_natbib}

\newpage
\appendix
\input{sections/appendix}

\end{document}

%% file: sections/01introduction.tex
\section{Introduction}

Nowadays, dialogue systems are ubiquitous in customer service and voice-based assistants.
One of the main uses of this technology is supporting humans in accomplishing tasks that might require accessing and navigating large knowledge bases (e.g., movies search).
A dialogue system architecture is typically composed of a natural language understanding (NLU) module, a dialogue management (DM) module, and a natural language generation (NLG) module \cite{DBLP:books/lib/JurafskyM09,williams2016dialog}.
First, the NLU component extracts a meaning representation from the user utterance based on which the DM generates the next system action by reasoning over the meaning representation and communicating with external applications if necessary. For example, the DM may retrieve information from external knowledge graphs (KG) to answer the user's query based on the dialogue history.
This process requires the DM to convert the output of NLU to a query to be issued to the backend.
Given the difficulty of this step, which is often domain-dependent, the DM component might require the design of hand-crafted rules. However, such rules are usually not scalable to different applications. They could require considerable effort to cover all possible cases/dialogue flows, leading to expensive costs to design new applications.
Moreover, in several cases, users interacting with such assistants are forced to formulate specific queries in order to accomplish their objective, which might break user engagement.

To alleviate the problem of having to design expensive hand-crafted rules and breaking user experience, recent works have explored the possibility of building end-to-end dialogue systems \cite{wen2017network} and all-in-one response generation models \cite{serban2016building}.
\ure{Among them, since graph is one of the main structure to store knowledge, recent research~\citep{ghazvininejad2018knowledge,zhou2018commonsense,moon2019opendialkg,tuan2019dykgchat,yang2020graphdialog} has proposed methods to generate natural language responses according to both the dialogue history and external knowledge graph.}
Despite these innovative and inspiring methods, there are some shortcomings.
For instance, these methods are either not fully-interpretable or limited to small-scale knowledge graphs.

In this paper, we propose a novel dialogue differentiable knowledge graph model (DiffKG).
The DiffKG is a single transformer model that directly (1) generates a sequence of relations to perform multi-hop reasoning on a reified KG representation proposed by \cite{cohen2019scalable}, and then (2) generates responses using the retrieved entities.
To the best of our knowledge, this is the first dialogue model that can directly walk on a large-scale KG with flexibility and interpretability.
DiffKG allows having flexible entity values in the KG and handling novel entity values with an arbitrarily defined number of tokens.
The reasoning path of DiffKG consists of the predicted relations, thus allowing for transparency.

We run extensive experiments to \ure{test DiffKG performance on KG-grounded dialogues. We select Stanford Multi-domain Dialogues (SMD)~\cite{eric2017key} and propose a new dataset, SMD-Reasoning, to simulate scenarios requiring multiple reasoning types and select the OpenDialKG~\cite{moon2019opendialkg} to simulate scenarios requiring large-scale KG reasoning without preprocessing. We then compare DiffKG with state-of-the-art models on SMD and OpenDialKG and an additional baseline that flattens KGs into a textual form from which transformers can learn.}
Empirically, our experiments show that DiffKG can effectively be trained on large-scale KGs and demonstrate its robustness with modified triplets in a KG.
From the perspective of computation, DiffKG leads to relatively low extra time and memory usage compared to transformer models not using any KG information. 

In summary, our contributions are: 1) We propose DiffKG, a novel method that can effectively and flexibly incorporate large-scale KG; 2) We demonstrate that DiffKG is a model-agnostic method and can be applied to different model architectures; 3) We show that DiffKG is an interpretable method with low add-on latency at inference time.
\ure{Our code and processed datasets are released in \url{https://github.com/Pascalson/DiffKG-Dialog}.}

\begin{table*}[t!]\small
    \centering
    \begin{tabular}{c|c|l|l|c}\toprule[1pt]
        \multicolumn{3}{l|}{\bf Reasoning Type} & \bf Example & \bf Related Info. in KG\\\midrule[.5pt]
        
        \parbox[t]{6mm}{\multirow{8}{*}{\rotatebox[origin=c]{90}{\makecell{Semantic Form}}}} & \multicolumn{2}{l|}{\multirow{2}{*}{KG reasoning}}  & U: I need unleaded gas. &
        \multirow{2}{*}{
            \begin{minipage}{.17\textwidth}
                 \includegraphics[width=\linewidth]{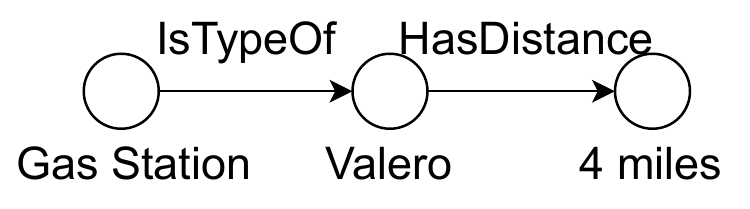}
            \end{minipage}
        }
        \\
        & \multicolumn{2}{l|}{} & R: inform Valero, 4 miles & \\\cmidrule[.5pt]{2-5}
        
        & \parbox[t]{6mm}{\multirow{6}{*}{\rotatebox[origin=c]{90}{\makecell{Logical\\Reasoning}}}} & \multirow{2}{*}{True/False} & U: Is it going to snow this week at Corona? & 
        \multirow{2}{*}{
            \begin{minipage}{.17\textwidth}
                 \includegraphics[width=\linewidth]{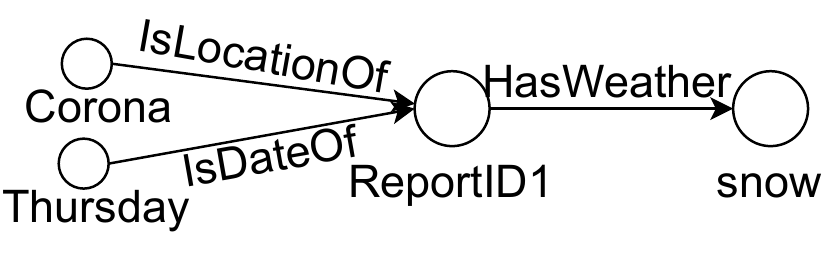}
            \end{minipage}
        }
        \\
        & & & R: Yes & \\\cmidrule[.5pt]{3-5}
        
        & & \multirow{2}{*}{Selection} & U: give me the direction to the nearest shopping mall. &
        \multirow{2}{*}{
            \begin{minipage}{.17\textwidth}
                 \includegraphics[width=\linewidth]{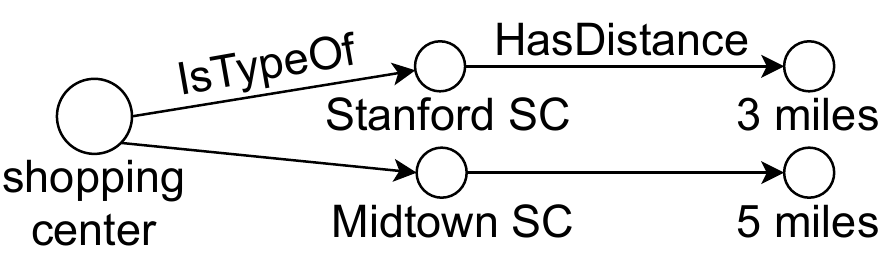}
            \end{minipage}
        }
        \\
        & & & R: inform Stanford Shopping Center, 3 miles & \\\cmidrule[.5pt]{3-5}
        
        & & \multirow{2}{*}{Extraction} & U: What gas stations are here? & \multirow{2}{*}{\makecell{No gas station\\in the available KG}}\\
        & & & R: include poi\_type gas station & \\\midrule[.5pt]
        
        \parbox[t]{6mm}{\multirow{2}{*}{\rotatebox[origin=c]{90}{\makecell{NL\\Form}}}} & \multicolumn{2}{l|}{\multirow{2}{*}{KG reasoning}}  & U: Have you listen to any of the singer Kesha's song? &
        \multirow{2}{*}{
            \begin{minipage}{.17\textwidth}
                 \includegraphics[width=\linewidth]{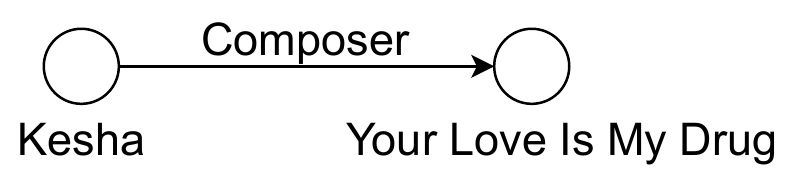}
            \end{minipage}
        }
        \\
        & \multicolumn{2}{l|}{} & R: I do enjoy in her music, especially ``Your Love Is My Drug'' &\\
        \bottomrule[1pt]
    \end{tabular}
    \caption{Example of different reasoning types and output formats (semantic and natural language forms) in a dialogue system with the related information in the accessible KGs.}
    \label{tab:task-examples}
\end{table*}

%% file: sections/02related_work.tex
\section{Related Work}

Recent years have seen a surge of new methods proposing end-to-end models that try to both understand natural language input text and search information. Two of the widely explored tasks are question-answering (QA) and dialogue generation.

\paragraph{QA.}
Multiple QA methods~\cite{weston2015towards, yin2016neural, hao2017end, rajpurkar2018know, verga2020facts, eisenschlos2021mate} have been proposed to tackle tasks that go beyond what is explicitly stated in the linguistic context~\cite{storks2019recent}.
For example, the benchmarks \cite{mihaylov2018can,reddy2019coqa,khot2020qasc,lin2021differentiable} are particularly useful for the model to extract information from external knowledge bases to answer questions.
Nonetheless, these studies mostly take the retrieved information from KG as the answer to a single question, while in dialogue we have to formulate an informative response to multi-turn dialogue history.

\paragraph{Dialogue Generation.}
Recent works have investigated the \ure{grounded} dialogue generation. These methods can be divided into three main categories.
First, \citet{dinan2018wizard,zhao2019low,tuan2020knowledge,kim2020beyond} extract useful knowledge from unstructured data to generate responses, such as information contained in passages and speaker's profiles.
Second, \citet{sordoni2015neural, long2017knowledge, zhu2017flexible, ghazvininejad2018knowledge, zhou2018commonsense,velivckovic2017graph,joshi2020dialograph,NEURIPS2020_e9462095,wang2021incorporating} utilize information from knowledge bases (either graphs or tables) to enhance the dialogue system. They usually train the entities and relations embeddings of the knowledge bases and incorporate these embeddings into the input representation to predict the response.
Third, \citet{moon2019opendialkg,tuan2019dykgchat,jung2020attnio} formulate the reasoning process more explicitly, as a path traversal over knowledge graphs. These methods further improve the transparency and explainability of the conversational agent and share the most similar idea with us.
However, they either only predict the reasoning path without generating responses or need subgraph sampling to reduce the scale of KG.
In this work, our approach uses a transformer model to jointly predict explicit reasoning paths over a large-scale knowledge graph and generate dialogue responses based on the reasoning results.

%% file: sections/03background.tex
\section{Background}

\subsection{Knowledge Graph for Dialogue System}
We assume that the knowledge of the system can be represented by a knowledge graph (KG) $\mathcal{G}=\{\mathcal{E},\mathcal{R}\}$, where $\mathcal{E}$ denotes the entities and $\mathcal{R}$ denotes the relations.
The knowledge graph $\mathcal{G}$ contains multiple triples describing the connections among entities and relations.
We denote the $k$-th triple of this graph as $(e_{k}^{h},r_{k},e_{k}^{t})$ , where $e_{k}^{h}$, $r_{k}$, $e_{k}^{t}$ are respectively the head entity, relation, and tail entity.
The total numbers of triples, entities, and relations are denoted as $N_\mathcal{T}$, $N_\mathcal{E}$, $N_\mathcal{R}$, respectively.\footnote{An example of the triples in $\mathcal{G}$ is a triple $e_{k}^{h}= \text{gas station}$, $r_k=\text{IsTypeOf}$, and $e_k^{t}=\textrm{Chevron}$. That is, ``gas station is the type of Chevron'' to this system. }


\subsection{Response Generation in Dialogue System}
If we define the dialogue history as a sequence of tokens that occurred during the user and system interactions, then a flattened dialogue history can be written as:
\begin{align}
\mathbf{x} = (x_1, x_2,..., x_m,..., x_M)
\end{align}
where $x_m$ is the $m$-th token in the dialogue history with $M$ tokens.
In an end-to-end dialogue system, we assume a dialogue system parameterized by $\theta$ exists that can predict a probability distribution of responses $P_\theta(\cdot|\mathbf{x}, \mathcal{G})$. The generated responses are sampled from this probability distribution.

\begin{figure*}
    \centering
    \includegraphics[width=.75\linewidth]{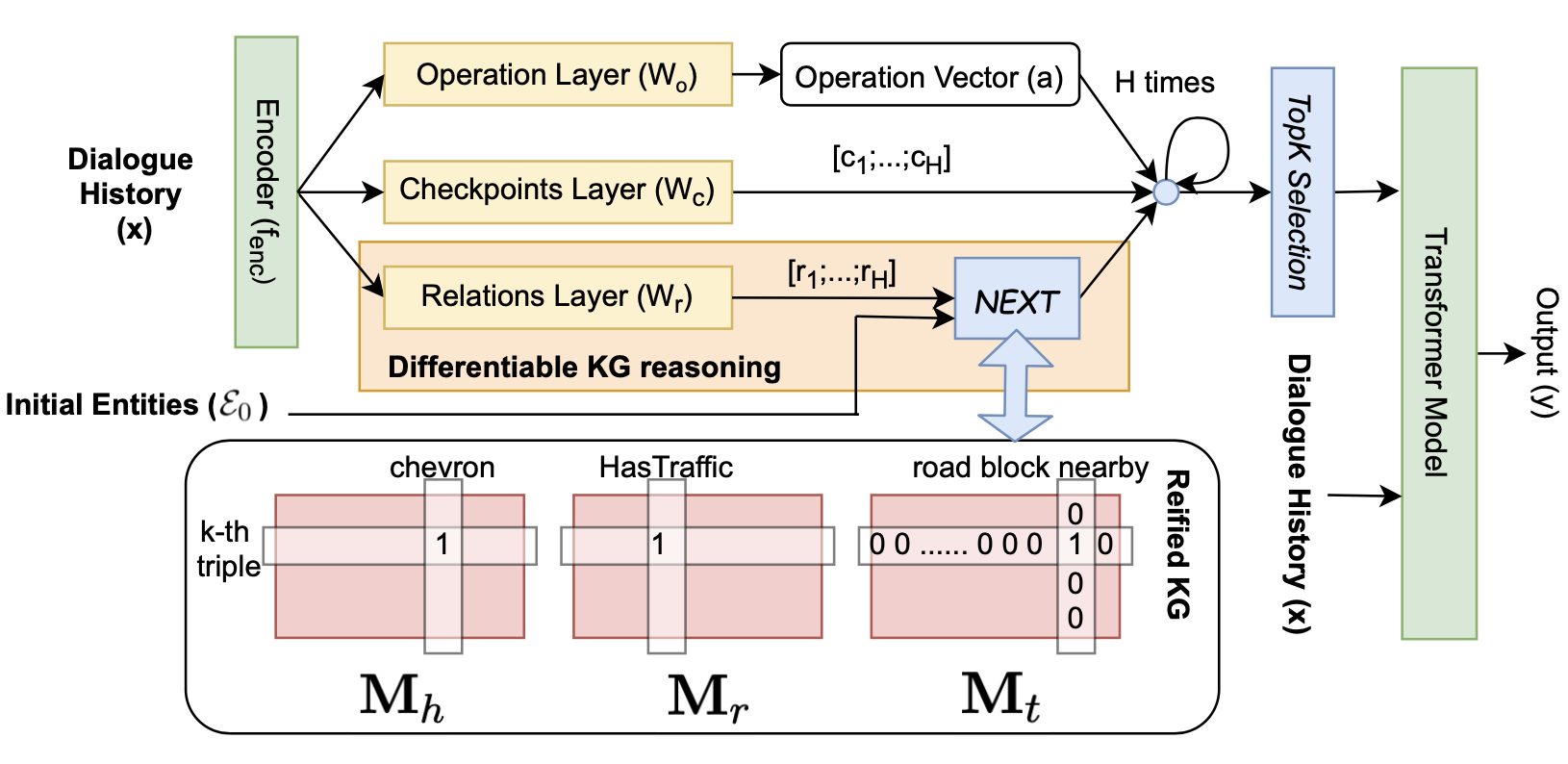}
    \caption{The illustration of proposed DiffKG, which leverages a pretrained transformer model (T5 or GPT2) and the Reified KG. The model generates the response depending on the predicted relation sequence $[r_1;...;r_H]$, thus being fully interpretable in terms of the used reasoning path.}
    \label{fig:diffkg-model}
\end{figure*}

\section{Problem Statement} 
\label{problem_statement}

We focus on understanding the ability of language models in performing reasoning during a conversation.
We consider two tasks that are usually required in dialogue scenarios and call them semantic form and natural language (NL) form in Table~\ref{tab:task-examples}.
First, given a dialogue history and a user's query, the task of semantic form is to predict the next system action, corresponding to the output of the DM module, based on the available knowledge.
In this case, we assume the expected output is the essential knowledge for an NLG module.
We argue that this task could help better evaluate if the response is correct or not and which type of reasoning can be more successfully handled.
Second, given a dialogue history and a user's query, the task of the NL form could be to directly output the response given by the system.
This setting with annotated reasoning path can shed light on understanding if the model can learn to support chit-chat and reasoning at the same time.

Moreover, we aim to understand models' reasoning capability both in the form of logical reasoning and over the provided knowledge. As illustrated in Table~\ref{tab:task-examples}, by KG reasoning, we refer to the ability of the model to retrieve information from an arbitrary scaled KG in multiple hops. Meanwhile, we refer to logical reasoning as the ability of the model to conduct operations such as evaluating whether a statement is true or false, selecting min/max from a list of alternatives, and extracting constraints.


We formulate the task that we focus on as follows: given the dialogue history $\mathbf{x}$ and currently accessible KG $\mathcal{G}$, can we extend a transformer model to predict a correct response $y$ in either semantic or NL form?
As illustrated in Table~\ref{tab:task-examples}, this task not only requires the model to accurately retrieve information from the KG, but also needs to do further logical operations on the information.
To solve this task, a model should also be able to effectively integrate the dialogue history $\mathbf{x}$ with the KG $\mathcal{G}$.

%% file: sections/04method.tex
\section{Proposed Approach}

Figure~\ref{fig:diffkg-model} illustrates our proposed architecture which contains four main parts: a dialogue history encoder, a differentiable KG reasoning module, a learnable logical operations module, and a response decoder (the transformer model).
Note that we experiment with two types of transformers: a causal language model GPT2~\cite{radford2019language} and an encoder-decoder model T5~\cite{raffel2020exploring}.
For GPT2, we reuse the same encoder that is used at the beginning of the process, i.e., $f_{enc}$ in Figure 1, as the final transformer that generates the response token by token. 
For T5, we reuse the same encoder as the encoder of the final transformer with a separate decoder that generates the response. 
Therefore, this method contains a single transformer model.
In following sections we present each module in detail.

\subsection{Dialogue History Encoder}
We use encoder model to project $\mathbf{x}$ and obtain the dialogue history embedding through $\Tilde{\mathbf{x}}=f_{enc}(\mathbf{x})\in {\rm I\!R}^{d}$, where $d$ is the hidden size of the encoder.
The embedding $\Tilde{\mathbf{x}}$ is first fed into an operation layer with parameters $\mathbf{W}_{o} \in {\rm I\!R}^{d\times d}$. The operation layer predicts the operation vector $\mathbf{a} = \mathbf{W}_o^T \Tilde{\mathbf{x}} \in {\rm I\!R}^{d}$.
At the same time, the embedding $\Tilde{\mathbf{x}}$ is also fed into a relation layer with parameters $\mathbf{W}_r \in {\rm I\!R}^{d\times N_\mathcal{R} H}$. The relation layer predicts the concatenation of a sequence of relations \ure{$\mathbf{r} = \{\mathbf{r}_h | 1\le h \le H\}$, where $\mathbf{r}_h \in {\rm I\!R}^{N_\mathcal{R}}$} is the relation to be used at the $h$-th hop in the programmed walking block and $H$ is the maximum number of hops.
The embedding $\Tilde{\mathbf{x}}$ is also fed into a checkpoints layer with parameters $\mathbf{W}_c \in {\rm I\!R}^{d\times 2H}$. This layer produces the concatenation of a sequence of walk-or-check vectors $ \mathbf{c} = \{c_h |  1\le h \le H\}$, where $c_h \in {\rm I\!R}^2$ is the walk-or-check vector at the $h$-th hop to determine the weights of the programmed walking module and the operation vector.
\begin{equation}
\begin{split}
    &  \Tilde{\mathbf{x}} = f_{enc}(\mathbf{x})\,,\\
    & \mathbf{a} = \mathbf{W}_o^T \Tilde{\mathbf{x}}\,,\\
    & \mathbf{r} =\mathbf{W}_r^T \Tilde{\mathbf{x}}\,,\\
    & \mathbf{c} = \texttt{softmax}(\mathbf{W}_c ^T \Tilde{\mathbf{x}})\,.\\
\end{split}
\end{equation}

\subsection{Differential Knowledge Graph Reasoning}

To ensure that our model can scale to larger KGs, we adopt the reified KG representation proposed by~\cite{cohen2019scalable}.
The reified KG represents the graph $\mathcal{G}$ using three sparse matrices: \ure{head matrix $\mathbf{M}_h\in {\rm I\!R}^{N_\mathcal{T}\times N_\mathcal{E}}$, relation matrix $\mathbf{M}_r\in {\rm I\!R}^{N_\mathcal{T}\times N_\mathcal{R}}$, and tail matrix $\mathbf{M}_t\in {\rm I\!R}^{N_\mathcal{T}\times N_\mathcal{E}}$}.
An entry $(i, e)$ in $\mathbf{M}_h$ or $\mathbf{M}_t$ with value $1$ indicates that the $i$-th triple in the KG has entity $e$ as the head or the tail; an entry $(i, r)$ in $\mathbf{M}_r$ with value $1$ indicates that the $i$-th triple in the knowledge graph has the relation $r$. Since often in practical settings most entries in the three matrices are zero, saving them into sparse matrices can significantly reduce memory consumption \cite{cohen2019scalable}.


After predicting the relation sequence \ure{$\mathbf{r}$}, we start the graph traversal from a given set of initial entities \ure{$\mathcal{E}_0\subseteq \mathcal{E}$}.
We first map the initial entities into a vector $\mathbf{e}_1 = [\mathbbm{1}(e\in \mathcal{E}_0), \forall e \in \mathcal{E}]$. That is, each entry of $\mathbf{e}_1\in {\rm I\!R}^{N_\mathcal{E}}$ has value 1 if that entity is in the initial entities list $\mathcal{E}_0$, otherwise, the entry is zero.  We then predict the next (temporary) entity vector $\mathbf{e}_2$ by conducting a $\texttt{Next}$ module:
\ure{
\begin{equation}
    \mathbf{e}^r_{h+1} = \texttt{Next}(\mathbf{e}_h, \mathbf{r}_h)\,,
\end{equation}
}
where 
\begin{equation}
    \texttt{Next}(\mathbf{e}_h,\mathbf{r}_h)= \frac{\mathbf{M}_t^T (\mathbf{M}_h \mathbf{e}_h \odot \mathbf{M}_r \mathbf{r}_h)}{||\mathbf{M}_t^T (\mathbf{M}_h \mathbf{e}_h \odot \mathbf{M}_r \mathbf{r}_h)||_2+\epsilon} \,,
\end{equation}
Here $\epsilon$ is an arbitrary small number to offset the denominator and prevent division by zero. We introduce the normalized $\texttt{Next}$ to solve the issue with the method proposed by~\cite{cohen2019scalable} for knowledge graph completion defined as $\texttt{Follow}(\mathbf{e}_h,\mathbf{r}_h) = \mathbf{M}_t^T (\mathbf{M}_h \mathbf{e}_h \odot \mathbf{M}_r \mathbf{r}_h)$; since in a dialogue model, we can seldom predict the relation vectors that perfectly match the entity vectors. That is, if directly using the $\texttt{Follow}$ module in~\cite{cohen2019scalable}, the $||\mathbf{e}_h||_2$ will not be one and will vanish as the hop number $h$ increases. 
Specifically, note that in our proposed module, the predicted relations $\mathbf{r}_h$ are independent of the traversed entities $\mathbf{e}_h$. For instance, finding the ``distance'' of ``the nearby gas station'' is independent of whether the nearby gas station is ``Chevron'' or ``Shell''.

To allow the model to dynamically select the number of reasoning hops, we add a relation type ``ToSelf'' into $\mathcal{R}$ and connect each entity to itself by ``ToSelf''. More specifically, the KG will contain triples $(e^h_k,r_k,e^t_k)$ for all $e^h_k=e^t_k\in \mathcal{E}$ and $r_k=\texttt{ToSelf}$. 


\subsection{Entity Embeddings}
At each hop, we further conduct the operation vectors $\mathbf{a}$ on the entities weighted by the entity vector $\mathbf{e}_h$.
First, we tokenize each entity and represent it by the concatenation of its token embeddings.
This step allows (1) representing entities with longer texts such as phrases and sentences, and (2) eliminating the effort to retrain entity embeddings whenever new entity values are added. 
The entity embeddings can then be represented as a tensor $\mathbf{E}\in {\rm I\!R}^{N_\mathcal{E}\times d \times m}$, where $m$ is the maximum number of tokens of entities\footnote{\ure{In our experiments, we compute the maximum length of all entities and pad shorter entities to the length of $m$.}}.

\subsection{Learnable Logical Operation and Checkpoints}
We compute the transformed entity embeddings by element-wise multiplication of the entity embeddings $\mathbf{E}$ with the entity vector $\mathbf{e}_h$ at the $h$-th hop. Next, the dot product of the operation vectors and the transformed entity embeddings is passed to a softmax layer as the entity vector at the next hop: 
\begin{equation}
    \mathbf{e}^a_{h+1} = \texttt{softmax}\left(\mathbf{a}(\mathbf{E} \odot \mathbf{e}_h)\right)\,,
\end{equation}

Further, at the $h$-th hop we use the walk-or-check vector $\mathbf{c}_h$ to combine the $\texttt{Next}$ and operation modules above. The combined entity vector is given by:
\begin{equation}
  \begin{split}
    \mathbf{e}_{h+1} & = \mathbf{c}_h^T
    \left [\begin{array}{c}
    \mathbf{e}^r_{h+1}\\
    \mathbf{e}^a_{h+1}
    \end{array}\right ] \\
    & = \mathbf{c}_h^T
    \left [\begin{array}{c}
    \texttt{Next}(\mathbf{e}_h,\mathbf{r}_h)\\
    \texttt{softmax}\left(\mathbf{a}(\mathbf{E} \odot \mathbf{e}_h)\right)
    \end{array}\right ]\,,
  \end{split}
\end{equation}

\subsection{Response Decoder}
After $H$ hops reasoning is done, the entities with top-$k$ values in the entity vector $\mathbf{e}_H$ are selected, indicating that they have the highest probability to be retrieved from the graph.
These entities are converted into their embeddings in $E$ and multiplied by their values in $\mathbf{e}_H$. These entity embeddings are then concatenated with the dialogue history $\mathbf{x}$. The concatenated vectors are fed as the input into the transformer model to predict the response token by token. The predicted probability distribution over the output space can be written as $P(\cdot|\mathbf{x},\mathbf{M}_h,\mathbf{M}_r,\mathbf{M}_t)$.
Since all components are differentiable, all modules can be trained end-to-end with the dialogue history $\mathbf{x}$ and the reified KG representation $\{\mathbf{M}_h,\mathbf{M}_r,\mathbf{M}_t\}$ using the cross-entropy loss with the ground-truth output $y$ as the labels.
\begin{equation}
    L = \sum_{(\mathbf{x},y)} - \log P(y | \mathbf{x},\mathbf{M}_h,\mathbf{M}_r,\mathbf{M}_t)\,.
\end{equation}

\ure{During the inference time, the reasoning modules (relation layer, operation layer, and checkpoints layer) work exactly the same as the training stage, the only difference is that the response decoder is fed with predicted tokens in previous time steps (inference stage) instead of the ground-truth output (training stage).}

%% file: sections/05experiments.tex
\section{Experiments}

\subsection{Datasets}
We evaluate our proposed approach on \ure{three datasets. Among them, we use Stanford Multi-domain Dialogues (SMD)~\cite{eric2017key} and OpenDialKG~\cite{moon2019opendialkg} to test} the methods generalizability on different dialogue types (task-oriented / chit-chat) and scales of structured knowledge (pairwise database / universal KG).
\ure{To further analyze the reasoning ability, we propose a new dataset, SMD-Reasoning, by modifying the output of SMD dataset from natural language responses to actions paired with their reasoning types.}

\paragraph{Stanford Multi-domain Dialogues (SMD)}
The SMD dataset~\cite{eric2017key} is composed of two-speaker conversations, 
where a driver talks with the car assistant to tackle tasks in three domains: scheduling, navigation, and weather forecasting.
Each dialogue focuses on one domain and is paired with a database having the related information.
We convert the original database into two formats: (1) the natural language descriptions (NLD) and (2) the KG.
The NLD form allows us to investigate the ability of the model to interpret unstructured knowledge, while the KG form could be a more extensible structured knowledge compared to tables.

\paragraph{OpenDialKG}
OpenDialKG dataset~\cite{moon2019opendialkg} is composed of two-speakers recommendation and chit-chat style conversations.
Each turn in a dialogue is annotated with the reasoning path on the provided KG, which is filtered from Freebase~\cite{bollacker2008freebase}.
The resulting KG has 1,190,658 triples, 100,813 entities and 1,358 relations.
We randomly \ure{split 70/15/15\%} for train/valid/test sets as described in \cite{moon2019opendialkg, jung2020attnio} since they do not release their splits.

\paragraph{SMD-Reasoning}
To make SMD dataset suitable for more precise evaluation of reasoning abilities, we \ure{manually label and} convert it to the SMD-Reasoning dataset.
We first remove the natural language part from the original responses and only leave the action word (e.g., inform) along with the information being conveyed.
We divide the dataset into three main reasoning types: informing items, selecting min/max, and evaluating true/false.
To validate if the models can identify whether the needed knowledge is in the database, we add a new reasoning type for extracting constraints, by removing the needed knowledge from the database and changing the output to ``include [knowledge description]'' as shown in Table~\ref{tab:task-examples}.
See Appendix~\ref{appx:smd-reasoning-stats},\ref{appx:smd-kg} for statistics of these datasets. 

\subsection{Evaluation Metrics}
We use different evaluation methods for the three datasets.
For SMD, we follow prior work~\cite{yang2020graphdialog} and use BLEU~\cite{papineni2002bleu}
, and Entity F1 scores on each domain. 
For OpenDialKG, we follow the descriptions in prior works~\cite{moon2019opendialkg,jung2020attnio} to evaluate the path@k scores, i.e., if the ground-truth path is ranked top-k in the predicted paths probabilities.
Moreover, since our method not only can predict the reasoning path as prior works but also can predict the response, we also use the BLEU score to get the approximated evaluation of the response quality compared to ground-truth. Note that prior work has discussed that BLEU scores may not match human intuition~\cite{liu2016not}, but we use them here as an approximated evaluation for reference.

For SMD-Reasoning, the output is more deterministic and does not include diverse sentence structures.
Therefore, we compute the F1 score and the {\it exact match (EM)} score of prediction and the ground-truth.
The EM score is calculated by removing the order of the prediction since the labels of SMD-Reasoning dataset follow the order of knowledge description appearing in the original ground-truth responses and may not have the same order as generated outputs.
The EM score can be written as:
\begin{equation}
    \texttt{EM} =  \frac{1}{T} \sum \mathbbm{1}(\texttt{sort}(\hat{y}) = \texttt{sort}(y))\,.
\end{equation}
where $\hat{y}$ is inferred from the model using argmax sampling and $T$ is total number of examples.


\subsection{Implementation Details}
Since the proposed method is model-agnostic, we implement it on GPT2~\cite{radford2019language} 
and T5~\cite{raffel2020exploring}. 
Specifically for the T5 model, we use the unifiedQA-T5 model~\cite{khashabi2020unifiedqa} which is pretrained on question answering tasks that also need to do reasoning.
However, we empirically find that T5 generally has better performance than GPT2, thus using T5 model in most experiments.
\ure{For OpenDialKG, since the ground-truth relations exist, we take them as an additional supervision signal as~\cite{moon2019opendialkg}.
Also, since we observe that there is only KG reasoning type in OpenDialKG, we do not use operation layer and checkpoints layer for the dataset.}
The hyperparameter settings are in Appendix~\ref{appx:exp-details}.

\subsection{Baselines}
We compare our proposed DiffKG model with the state-of-the-art models on OpenDialKG reported in~\cite{moon2019opendialkg,jung2020attnio} and the state-of-the-art graph-based model on SMD~\cite{yang2020graphdialog, DBLP:conf/emnlp/GouLLDS21} with their reported baselines including sequence-to-sequence models with and without attention (S2S and S2S+Attn)~\cite{luong2015effective}, pointer to unknown (Ptr-Unk)~\cite{gulcehre2016pointing}, GraphLSTM~\cite{peng2017cross}, BERT~\cite{devlin2019bert}, Mem2Seq~\cite{madotto2018mem2seq} and GLMP~\cite{wu2018globaltolocal}.
We follow their metrics and train our model on their preprocessed data for fair comparisons.
To further analyze the reasoning ability, we propose two more baselines based on different ways of leveraging pretrained language models.
(1) {\bf NoInfo} model does not take any format of knowledge as the input, aiming to test the performance of a fine-tuned vanilla transformer model on each dataset.
(2) {\bf FlatInfo} model constructs the input by concatenating the dialogue history with the NLD form of knowledge as~\cite{beygi2022logical}, allowing us to investigate the ability of the model to interpret unstructured knowledge.

\begin{table}[t]\small
    \centering
    \begin{tabular}{l|c|c|ccc}\toprule[1pt]
        \multirow{2}{*}{Model} & \multirow{2}{*}{BLEU} & \multicolumn{4}{c}{Entity F1}\\
        & & All & Sche. & Wea. & Nav.\\\midrule[0.5pt]
        S2S & 8.4 & 10.3 & 9.7 & 14.1 & 7.0\\
        S2S+Attn & 9.3 & 19.9 & 23.4 & 25.6 & 10.8\\
        Ptr-Unk & 8.3 & 22.7 & 26.9 & 26.7 & 14.9\\
        GraphLSTM & 10.3 & 50.8 & 69.9 & 46.6 & 43.2\\
        BERT & 9.13 & 49.6 & 57.4 & 47.5 & 46.8\\
        Mem2Seq & 12.6 & 33.4 & 49.3 & 32.8 & 20.0\\
        GLMP & 12.2 & 55.1 & 67.3 & 54.1 & 48.4\\
        GraphDialog & 13.7 & 57.4 & 71.9 & 59.7 & 48.6\\
        COMET-graph & 14.4 & 56.7 & 71.6 & 48.7 & 50.4\\\midrule[0.5pt]
        T5-DiffKG & 16.04 & 56.2 & 67.2 & 61.5 & 46.7\\\bottomrule[1pt]
    \end{tabular}
    \caption{The results on SMD dataset. S2S, S2S+Attn, Ptr-Unk, GraphLSTM, BERT, Mem2Seq, GLMP, GraphDialog are reported from \cite{yang2020graphdialog} and \ure{COMET-graph from~\cite{DBLP:conf/emnlp/GouLLDS21}}. Our DiffKG achieves the highest BLEU and comparable F1 scores with baselines.}
    \label{tab:smd-graphdialog-results}
\end{table}

\begin{table}[t]\small
    \centering
    \begin{tabular}{l|cccc}\toprule[1pt]
        Model & path@1 & path@5 & path@10 & BLEU\\\midrule[.5pt]
        Seq2Seq & 3.1 & 29.7 & 44.1 & - \\
        Tri-LSTM & 3.2 & 22.6 & 36.3 & - \\
        EXT-ED & 1.9 & 9.0 & 13.3 & - \\
        DialKG & 13.2 & 35.3 & 47.9 & - \\
        Seq2Path & 14.92 & 31.1 & 38.68 & - \\
        AttnFlow & 17.37 & 30.68 & 39.48 & - \\
        AttnIO-AS & 23.72 & 43.57 & 52.17 & - \\\midrule[0.5pt]
        T5-NoInfo & - & - & - & 14.51 \\
        T5-DiffKG & 26.80 & 54.33 & 61.75 & 15.37 \\
        \bottomrule[1pt]
    \end{tabular}
    \caption{The results on OpenDialKG dataset. The four baselines from Seq2Seq to DialKG Walker are reported from \cite{moon2019opendialkg} and the other three baselines from Seq2Path to AttnIO-AS are reported from \cite{jung2020attnio}. Our DiffKG achieves the highest path@k scores and is the only one that can simultaneously generate responses.}
    \label{tab:opendialkg-results}
\end{table}

\subsection{Results}
The results on SMD and OpenDialKG are shown in Table~\ref{tab:smd-graphdialog-results} and Table~\ref{tab:opendialkg-results}.
On SMD dataset, we observe that DiffKG outperforms the baselines on BLEU by 11.4\% (relative change of 16.04 and 14.4) and achieves comparable entity F1 scores with GLMP, GraphDialog and COMET-graph.
DiffKG might not improve the entity F1 scores because that prior works group the text inside an entity together (e.g., ``road block nearby'' becomes a single word ``road\_block\_nearby'' in vocabularies).
In contrast, we use a universal tokenizer so as to prevent heavy preprocessing and specialized vocabularies.
This means that DiffKG can perform similarly with state-of-the-art to retrieve knowledge without a tokenizer specified for each dataset.
On OpenDialKG dataset, we observe that DiffKG outperforms the baselines in terms of path@k scores and can simultaneously outperform T5 in terms of Entity F1 and BLEU.
These demonstrate that DiffKG can retrieve accurate paths for reasoning and effectively incorporate reasoning into response generation.

\begin{table}[t]\small
    \centering
    \begin{tabular}{c|l|cc}\toprule[1pt]
        \bf Test KG & \bf Method & \bf EM & \bf F1 \\\midrule[.5pt]
        \multirow{6}{*}{Fixed} & GPT2-NoInfo & 10.71 & 43.78\\
        & GPT2-FlatInfo & 14.08 & 47.57\\
        & GPT2-DiffKG & 16.39 & 51.06\\
        & T5-NoInfo & 10.50 & 44.27\\
        & T5-FlatInfo & 28.99 & 66.15\\
        & T5-DiffKG & 27.52 & 63.93\\\midrule[0.5pt]
        \multirow{2}{*}{Shuffled} & T5-FlatInfo & 17.02 & 54.51\\
        & T5-DiffKG & 27.52 & 64.00\\
        \bottomrule[1pt]
    \end{tabular}
    \caption{The results on SMD-Reasoning dataset.}
    \label{tab:smd-reasoning-results}
\end{table}

We also investigate the results of SMD-Reasoning dataset as shown in Table~\ref{tab:smd-reasoning-results}.
We find that DiffKG improves NoInfo by 16.6\% and 44.4\% F1 scores respectively on GPT2 and T5 models.
This demonstrates that DiffKG can utilize knowledge effectively to improve the generation without access to information.
In contrast, although FlatInfo gives similar performances as DiffKG on the SMD-Reasoning dataset, it cannot be run on OpenDialKG due to computational costs. More specifically, FlatInfo requires the knowledge \ure{graph} to be transformed into sentences, which will result in at least a million tokens as the model inputs for OpenDialKG \ure{(since the number of triples is a million without designed subgraph sampling)}, which is not a practical number.

%% file: sections/06analysis.tex
\begin{table*}[t]\small
  \begin{center}
    \begin{tabular}{l|cc|cc|cc|cc|cc|cc|cc}\toprule[1pt]
    \multirow{3}{*}{\bf Method} & \multicolumn{6}{c|}{\bf Domains} & \multicolumn{8}{c}{\bf Reasoning Types}\\\cmidrule[.5pt]{2-15}
    &\multicolumn{2}{c|}{\bf Schedule} &\multicolumn{2}{c|}{\bf Navigation} &\multicolumn{2}{c|}{\bf Weather}
    &\multicolumn{2}{c|}{\bf Inform} &\multicolumn{2}{c|}{\bf Selection} &\multicolumn{2}{c|}{\bf Extraction} &\multicolumn{2}{c}{\bf True/False}\\
     & EM & F1 & EM & F1 & EM & F1
     & EM & F1 & EM & F1 & EM & F1 & EM & F1\\\midrule[0.5pt] 
    GPT2-NoInfo & 3.49 & 45.7 & 4.63 & 41.6 & 27.5 & 46.8
    & 5.03 & 45.2 & 1.45 & 47.4 & 3.06 & 24.0 & 68.6 & 68.6\\
    GPT2-DiffKG & 9.30 & 53.0 & 9.65 & 47.6 & 34.4 & 56.5
    & 8.04 & 50.8 & 0.00 & 48.5 & 31.6 & 53.5 & 56.9 & 56.9\\
    T5-NoInfo & 0.00 & 44.6 & 4.63 & 40.9 & 29.0 & 50.7
    & 3.02 & 44.9 & 8.70 & 49.1 & 1.02 & 25.2 & 70.6 & 70.6\\
    T5-DiffKG & 20.9 & 63.8 & 19.3 & 61.9 & 48.1 & 68.1
    & 18.1 & 61.7 & 11.6 & 62.4 & 50.0 & 73.5 & 70.6 & 70.6\\\bottomrule[1pt]
    \end{tabular}
    \caption{Detailed Evaluation Results of SMD-Reasoning dataset}
    \label{tab:smd-types-results}
  \end{center}
\end{table*}

\begin{table*}[t]\small
    \centering
    \begin{tabular}{c|l}
        \toprule[1pt]
        \multirow{8}{*}{SMD-Reasoning} & User: check the date and time of my doctor's appointment\\
        & (Reasoning Path: Doctor Appointment $\xrightarrow[]{\text{HasDate, HasTime, ToSelf}}$ Tuesday, 11am, doctor appointment)\\
        & DiffKG: inform 11 am tuesday doctor appointment\\
        \cmidrule[0.5pt]{2-2}
        & User: Car I need to get to a gas station, please show me the nearest one\\
        & Assistant: There is Valero 7 miles away with moderate traffic on our way\\
        & User: Alright, where is it located?\\
	    & (Reasoning Path: Gas Station $\xrightarrow[]{\text{IsTypeOf}}$ Valero $\xrightarrow[]{\text{ HasAddress, ToSelf}}$ 200 Alester Ave, Valero)\\
        & DiffKG: inform 200 Alester Ave Valero\\
        \midrule[0.5pt]
        \multirow{3}{*}{OpenDialKG} & Speaker A: Do you have any info on Toni Kroos?\\
        & (Reasoning Path: Toni Kroos $\xrightarrow[]{\text{$\sim$Player Statistics}}$ Germany national football team)\\
        & DiffKG: Toni Kroos is German footballer who plays for the Germany national football team.\\
        \bottomrule[1pt]
    \end{tabular}
    \caption{Generated examples and the reasoning path.}
    \label{tab:examples}
\end{table*}

\begin{table*}[t]\small
    \centering
    \begin{tabular}{c|p{12cm}}
        \toprule[1pt]
        \bf Error Type & \bf Example\\\midrule[0.5pt]
        \multirow{4}{*}{\makecell{Unclear Information\\Requirement}} & User: What gas stations are here?\\
        & Ground-truth: inform Chevron\\
        & DiffKG: inform 5 miles, Chevron\\
        & (Our comment: Although this prediction is considered wrong for the EM metric,  the “5 miles” of DiffKG output is the correct distance from “Chevron” and might be needed in a good response.) \\
        \midrule[0.5pt]
        \multirow{4}{*}{\makecell{Incomplete Reasoning\\but Faithful Response}}& User: Where is the closest grocery store?\\
        & Ground-truth: Inform 4 miles, Whole Foods, Safeway.\\
        & DiffKG: inform 4 miles, grocery store, 819 Alma St, Whole Foods\\
        & (Our comment: The 4 miles, grocery store, 819 Alma St are all correct entities about Whole Foods. Nonetheless, this reasoning process neglects another grocery store Safeway which is also 4 miles away.) \\
        \midrule[0.5pt]
        \multirow{5}{*}{\makecell{Correct Reasoning\\but Wrong Response}} & Speaker A: Do you know Don Hall?\\
        & Ground-truth: Don Hall wrote the Princess and the Frog a romance story starring Jenifer Lewis. Do you like Romance?\\
        & Reasoning Path: Don Hall $\xrightarrow[]{\text{$\sim$written by}}$ The Princess and the Frog\\
        & DiffKG: Yes, he wrote The Little Dolls.\\
        & (Our comment: The reasoning path is correct to find out the script written by Don Hall. However, the generation process fails to properly utilize the retrieved entity.) \\
        \bottomrule[1pt]
    \end{tabular}
    \caption{The error analysis with three major error types across datasets.}
    \label{tab:error-analysis}
\end{table*}

\subsection{Quantitative Analysis}
\label{subsec:robustness}

To test the robustness of the methods towards accurately locating information, we shuffle the information order. This evaluation is to simulate the cases that extra information is arbitrarily added when deploying a dialogue system. Specifically, the order of the knowledge context for FlatInfo and the order of knowledge triples are changed during inference time.
As shown in the last two rows in Table~\ref{tab:smd-reasoning-results}, the performance of FlatInfo drops while DiffKG remains about the same.
This indicates that the slight superior performance of FlatInfo with the original order can come from the blackbox tricks to group the nearby knowledge in the inputs. When this implicit trick is broken down, the DiffKG shows much better robustness and performance.


To investigate the difficulty of each domain and reasoning type, we divide the results accordingly in Table~\ref{tab:smd-types-results}.
As presented in the domains part, the models achieve the highest EM and F1 on the weather domain.
We conjecture the reason is that the weather domain includes more reasoning types (weather:4, navigate:3, schedule:2 as in Appendix~\ref{appx:smd-reasoning-stats} Table~\ref{tab:SMD-Reasoning-stats}), thus reflecting more balanced reasoning ability.
In the reasoning types part, we observe that true/false is less well coped by DiffKG; however, DiffKG improves the extraction.
This shows that DiffKG can effectively check the existence of required knowledge and then query the database.


Regarding to the computational costs (on SMD-Reasoning dataset using T5 model), we found that DiffKG requires about 5.85GB memory during training and has 30ms inference latency. This could be an acceptable add-on memory usage and inference time compared to a model without knowledge reasoning (3.13GB; 30ms). Especially when a baseline like FlatInfo consumes much more (18.56GB; 50ms).

\subsection{Qualitative Analysis}

We visualize the generated examples and the symbolic reasoning path by DiffKG on SMD-Reasoning and OpenDialKG datasets in Table~\ref{tab:examples}.
The examples show that DiffKG can capture some naturally occurring phenomena in this dataset:
(1) the KG reasoning path can be 1 to multiple hops; (2) the reasoning will diffuse to multiple paths (e.g., DiffKG simultaneously applies ``HasDate'',``HasTime'',``ToSelf'' to ``Doctor Appointment'').
Along with analyses in previous subsections, we observe that DiffKG can extract interpretable reasoning paths and generate corresponding outputs using reasonable computational costs.

However, even though DiffKG can maintain or improve performance while doing interpretable reasoning on any scaled KG, errors might happen in some cases.
As listed in Table~\ref{tab:error-analysis}, we found that across the datasets, the three main error types of DiffKG are: (1) unclear information requirement in the dataset, (2) incomplete reasoning ability but faithful response generation, and (3) correct reasoning but hallucinated response prediction.
We argue that the first error type mainly comes from the mismatch among data points in the dataset and may not be able to be dealt with by models.
The second error type indicates that the KG reasoning module sometimes cannot retrieve all the needed information.
The third error type indicates that the module producing final output may not fully utilize the retrieved information.
These three points might provide a direction for further improvement.


%% file: sections/07conclusion.tex
\section{Conclusion and Future Work}

For a dialogue system, an effective reasoning method over structured databases is important.
In this work, we proposed DiffKG, an end-to-end model-agnostic method that does symbolic reasoning on any scale of KGs to enhance response generation.
Experiments demonstrated that using DiffKG, models are able to generate responses with interpretable KG reasoning paths at a modest extra cost.

This work can be extended in various ways.
While we solely consider efficient large-scale KG reasoning in dialogue generation, future work can incorporate domain fusion methods to consider the generalizability over domains or simultaneously use relation information.
Moreover, since DiffKG is a simple large-scale structured knowledge-empowered transformer with flexible entity values, future work can extend it to dialogue generation that needs to do table and text mixed reasoning and that needs to do both KG reasoning and other goals such as personalized dialogues, storytelling, etc.

%% file: sections/appendix.tex
\section{Dataset Details}
\label{appx:smd-reasoning-stats}
The statistics of datasets are in Table~\ref{tab:data-stats} and Table~\ref{tab:SMD-Reasoning-stats}. The OpenDialKG dataset is under CC-BY-NC-4.0 license. These datasets can be used for research purposes.

\begin{table}[h!]\small
    \centering
    \begin{tabular}{l|ccc}\toprule[1pt]
        Data & Train & Validation & Test \\\midrule[.5pt]
        SMD & 2425 & 302 & 304 \\
        OpenDialKG & 10971 & 2351 & 2351\\\bottomrule[1pt]
    \end{tabular}
    \caption{The number of dialogues in each data split.}
    \label{tab:data-stats}
\end{table}

\begin{table}[h!]\small
    \centering
    \begin{tabular}{l|ccc}\toprule[1pt]
         & Schedule & Navigation & Weather \\\midrule[.5pt]
        Inform & 364 & 1133 & 236 \\
        Selection & - & 686 & 39\\
        True/False & - & - & 543\\
        Extraction & 173 & 474 & 214\\\bottomrule[1pt]
    \end{tabular}
    \caption{The statistics of SMD Reasoning dataset with respect to domains and reasoning types. Note that since reasoning types are classified on turn-level, the total number in this table is larger than in Table~\ref{tab:data-stats} that counted on dialogue-level.}
    \label{tab:SMD-Reasoning-stats}
\end{table}

\section{SMD KG Construction}
\label{appx:smd-kg}

We write a simple, automatic program to construct KGs for SMD dataset mapped from the original annotated tables.

For the schedule and navigation domains in SMD, we directly map their table attributes to the relations $\mathcal{R}$ in our constructed KG. For the weather domain, we split each weather report into low temperature, high temperature, and weather. The resulting number of relations is 29, and the relations are listed in Table~\ref{tab:smd-relations}.
\begin{table}[h]
    \centering
    \begin{tabular}{l|p{5cm}}\toprule[1pt]
        Domain & Relations \\\midrule[0.5pt]
        Schedule & HasTime, HasDate, HasParty, HasRoom, HasAgenda, IsTimeOf, IsDateOf, IsPartyOf, IsRoomOf, IsAgendaOf\\
        Navigation & HasAddress, HasType, HasTraffic, HasDistance, IsAddressOf, IsTypeOf, IsTrafficOf, IsDistanceFrom\\
        Weather & IsEqualTo, HasLocation, HasWeather, HasLowTemp, HasHighTemp, HasDate, IsWeatherOf, IsLowTempOf, IsHighTempOf, IsLocationOf, IsDateOf\\
    \bottomrule[1pt]
    \end{tabular}
    \caption{The relations used in each domain in SMD dataset.}
    \label{tab:smd-relations}
\end{table}

In the schedule and navigation domain, each item in the original database with multiple attributes are transformed to KG triples as (event/point-of-interest, attribute, attribute value), e.g., \texttt{(tennis activity, HasTime, 7pm)} in schedule domain or \texttt{(Chevron, HasType, gas station)} in navigation domain.

In the weather domain, we add additional entities named ``ReportID\$digits\$'', where \$digits\$ will be replaced with an ID number.
Each item in the original database is in the format: (item, location, \$location), (item, \$date, \$weather\_report), where the \$weather\_report contains multiple information not simultaneously needed.
To make the KG of weather consistent with the KGs of schedule and navigation, we transform each item into (ReportID, location, \$location), (ReportID, HasDate, \$date), (ReportID, HasWeather, \$weather), (ReportID, HasLowTemp, \$low\_temparature), (ReportID, HasHighTemp, \$high\_temparature).

\section{Experiment Details}
\label{appx:exp-details}

The hyperparameters we set for DiffKG are $d$=the hidden size of the used pretrained trasnformer (T5-small: $d$=512; GPT2:$d$=768), $H$=5, max norm=$1.0$, batch size=$16$, and gradient accumulation steps=$2$ for at most $50$ epochs and train the model learning rate$\in \{5\times 10^{-5},6.25\times 10^{-5}\}$ (found that $6.25\times 10^{-5}$ is better) without learning rate decay. Our experiments were single runs with random initialization and were not further fine-tuned.

\section{Computational Cost Analysis}

\begin{figure}[t]
  \includegraphics[width=0.95\textwidth]{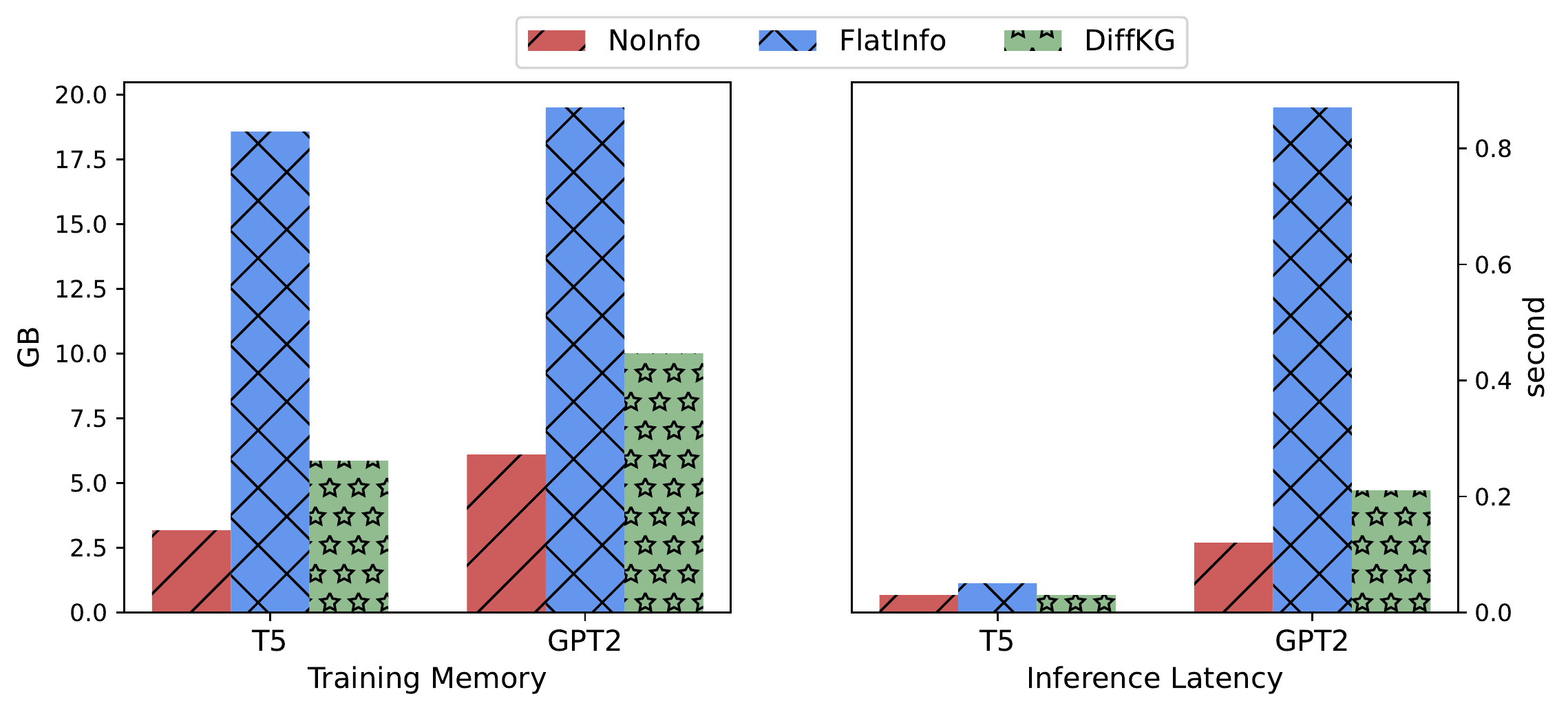}
  \caption{The comparison of the consumed training memory and inference latency.}
  \label{fig:computation}
\end{figure}

As plotted in Figure~\ref{fig:computation}, on SMD-Reasoning dataset, the consumed memory of FlatInfo is thrice the memory needed for DiffKG at training time, and its latency is about twice at inference time.
The difference in inference latency is even larger with GPT2 as the backbone model.
The reason is that the computational cost of a causal language model such as GPT2 largely depends on the input sequence length, which is one of the main issues of FlatInfo.

